\documentclass[conference]{IEEEtran}
\IEEEoverridecommandlockouts
\usepackage{cite}
\usepackage{amsmath,amssymb,amsfonts}
\usepackage{algorithmic}
\usepackage{graphicx}
\usepackage{textcomp}
\usepackage{xcolor}

\usepackage{tabularx}
\usepackage{multirow}
\usepackage{booktabs}
\usepackage{caption}

\usepackage{hyperref}
\hypersetup{pdfstartview=FitH,
            colorlinks=true,
            linkcolor=red,
            anchorcolor=blue,
            citecolor=black
            }

\def\BibTeX{{\rm B\kern-.05em{\sc i\kern-.025em b}\kern-.08em
    T\kern-.1667em\lower.7ex\hbox{E}\kern-.125emX}}
\begin{document}

\title{ES-Parkour:~Advanced Robot Parkour with Bio-inspired Event Camera and Spiking Neural Network}

\author{Qiang Zhang$^{1,2*}$, Jiahang Cao$^{1*}$, Jingkai Sun$^{1,2*}$, Yecheng Shao$^{3,4}$, Gang Han$^{2}$,  Wen Zhao$^{2}$, Yijie Guo$^{2}$, Renjing Xu$^{1\dagger}$
\thanks{$^\dagger$Corresponding author; $^*$Equal contribution.}
\thanks{$^{1}$The authors are with the Microelectronics Thrust, The Hong Kong University of Science and Technology (Guangzhou), Guangzhou, China. {\tt\small qzhang749@connect.hkust-gz.edu.cn, renjingxu@ust.hk}}%
\thanks{{$^{2}$The authors are with Beijing Innovation Center of Humanoid Robotics Co., Ltd.}
{\tt\small  jack.guo@x-humanoid.com}}
\thanks{$^{3}$The author is with Center for X-Mechanics, Zhejiang University, China.}
\thanks{$^{4}$The author is with Institute of Applied Mechanics, Zhejiang University, China.
 \tt\small shaoyecheng@zju.edu.cn  }%
}

\maketitle
\begin{abstract}

In recent years, quadruped robotics has advanced significantly, particularly in perception and motion control via reinforcement learning, enabling complex motions in challenging environments. Visual sensors like depth cameras enhance stability and robustness but face limitations, such as low operating frequencies relative to joint control and sensitivity to lighting, which hinder outdoor deployment. Additionally, deep neural networks in sensor and control systems increase computational demands.
To address these issues, we introduce spiking neural networks (SNNs) and event cameras to perform a challenging quadruped parkour task. Event cameras capture dynamic visual data, while SNNs efficiently process spike sequences, mimicking biological perception. Experimental results demonstrate that this approach significantly outperforms traditional models, achieving excellent parkour performance with just \textbf{11.7\%} of the energy consumption of an artificial neural network (ANN)-based model, yielding an \textbf{88.3\%} energy reduction. By integrating event cameras with SNNs, our work advances robotic reinforcement learning and opens new possibilities for applications in demanding environments.
\end{abstract}


Quadruped robotics has advanced significantly in both proprioceptive motion control ~\cite{reske2021imitation,hwangbo2019learning,wu2023learning,peng2020learning,iscen2018policies} and vision-based planning. These robots now perform a wide range of tasks in complex environments~\cite{zhuang2023robot,cheng2023extreme}, demonstrating potential applications in extreme conditions. However, challenges such as diverse lighting conditions, complex terrains, and energy efficiency remain underexplored.

Our study addresses these challenges by leveraging Spiking Neural Networks (SNNs) and event cameras to improve perception and motion control. Event cameras, which detect pixel-level changes at kilohertz frequencies and operate independently of lighting conditions, provide dynamic and high-frequency visual information. SNNs, inspired by biological neurons, process spike signals efficiently, reducing computational demands while maintaining performance. The integration of these technologies bridges the gap between fast joint control cycles and high-frequency perception, crucial for quadruped robots. Additionally, their low power consumption alleviates design trade-offs, reducing the need for bulky cooling systems and batteries.

\begin{figure*}[t!]
	\setlength{\tabcolsep}{1.0pt}
	\centering
	\begin{tabular}{c}
        \includegraphics[width=.8\textwidth]{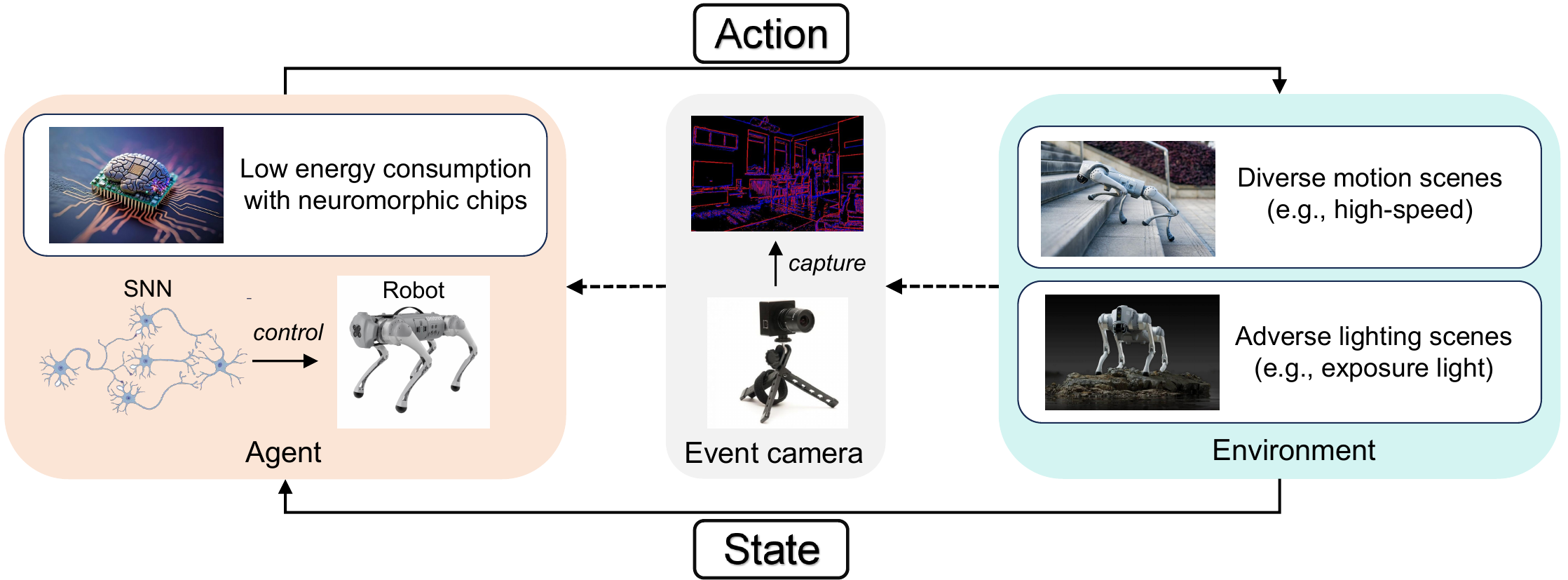}
	\end{tabular}
	\caption{\textbf{Pipeline of our bio-inspired reinforcement learning system.} Different from the previous standard vision-based robot system, our bio-inspired system is equipped with an event camera to capture event data from diverse scenes. The event is then processed by the spiking neural network which in turn dictates the robot's actions to the environment. The adoption of this brain-inspired approach yields three significant advantages: (1) enhanced stability in motion-intensive scenarios is achieved through the superior temporal resolution of the event data. (2) the system's resilience in fluctuating lighting conditions is ensured by the event camera's high dynamic range. (3) the inherently low energy consumption of the SNN contributes to the system's overall efficiency.
 }
	\label{fig:ideal_robot}
\end{figure*}

Using the Parkour task as a benchmark, we develop a simulation environment to rigorously evaluate the capabilities of quadruped robots under complex conditions. Unlike previous approaches that rely on 3D sensors such as LiDAR or depth cameras, our work demonstrates the potential of event cameras for effective perception in challenging environments. By combining SNNs and event cameras, we enhance efficiency, reduce computational costs, and advance the feasibility of real-world robotic applications. Figure~\ref{fig:ideal_robot} illustrates our bio-inspired system pipeline.

\begin{figure}[t]
	\setlength{\tabcolsep}{1.0pt}
	\centering
        \includegraphics[width=0.48\textwidth]{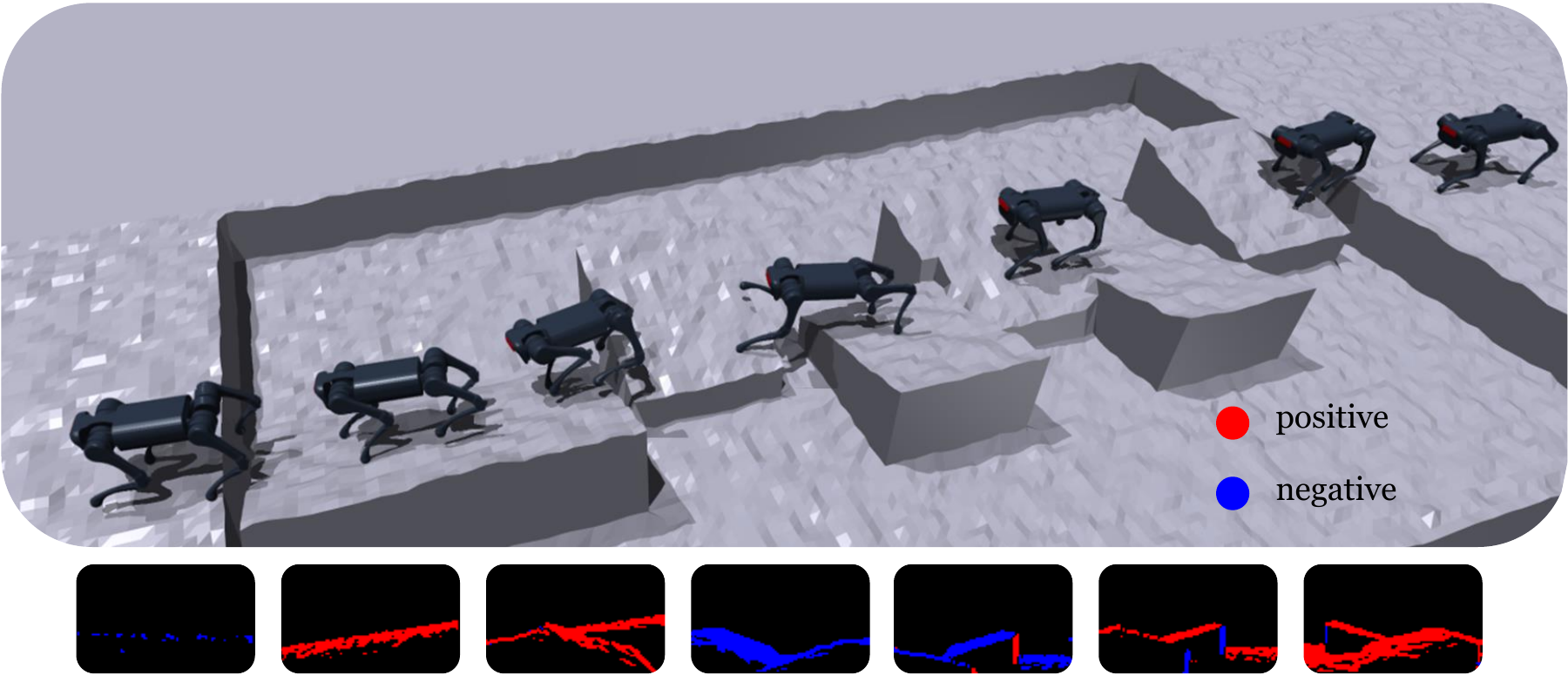}
	\caption{{\textbf{Demonstration of a quadruped robot performing parkour using spiking neural network under extreme lighting conditions.} The robot processes event images in real-time, where the red part and blue part denote the positive event and negative event, respectively.}}
	\label{fig:teaser}
    \vspace{-0.5cm}
\end{figure}

Our work makes three primary contributions:
\begin{enumerate}
\item We pioneer the implementation of a system-level design for quadruped robot parkour using SNNs and event cameras (in Figure~\ref{fig:teaser}), providing new insights into perception and motion control.
\item We successfully transition the end-to-end training of the quadruped robot RL network from ANNs to SNNs through a distillation method, significantly reducing computational burden and training complexity.
\item To our knowledge, this is the first demonstration of achieving complex quadruped robot control tasks across a variety of environments using a brain-inspired sensor. Our work ensures that robots possess robust perception and control capabilities in various environments, significantly broadening the application scope of brain-inspired devices in robots.
\end{enumerate}

Although breakthrough research has already been conducted in drones~\cite{davide2020dynamic} and autonomous driving~\cite{gehrig2024low}, the application of event cameras in legged robots remains limited. This scarcity is primarily due to the complexities involved in designing and implementing the systems. However, the combination of event cameras and spiking neural networks (SNN) holds significant potential for the robotics community. We hope our extensive simulation validation work will pave the way for further advancements in this field.

\section{Related Work}

\noindent\textbf{Legged Robot Agile Locomotion.}
Legged robots have advanced in agile locomotion through optimization-based~\cite{bledt2020extracting,di2018dynamic} and learning-based methods~\cite{reske2021imitation,hwangbo2019learning}. Trajectory optimization (TO) and Model Predictive Control (MPC) are commonly used for stable, dynamic motion but rely on detailed robot models and terrain estimation. Learning-based methods, in contrast, use neural networks to directly process visual data, avoiding complex modeling. Notable approaches include Rapid Motor Adaptation (RMA)\cite{kumar2021rma} and Adversarial Motion Priors (AMP)\cite{peng2021amp}, which generalize policies across terrains. Recent advances have integrated proprioception with vision, using depth sensors~\cite{imai2022vision} and 3D feature encoding~\cite{yang2023neural} for improved terrain perception.

\noindent\textbf{Robotic Parkour.}
Robotic parkour focuses on algorithms for navigating complex, high-risk terrains. Approaches like~\cite{zhuang2023robot} pre-train models under soft constraints and refine them with stricter ones. Hierarchical frameworks~\cite{hoeller2023anymal} combine motion and navigation policies for effective traversal. However, these methods mainly rely on traditional sensors like depth cameras and LiDAR, which struggle in challenging lighting and fast-moving scenes.

\noindent\textbf{Event Camera and SNN on robots.}
Event cameras capture high-speed, dynamic scenes asynchronously and are increasingly applied in robotics for tasks like Visual-Inertial Odometry (VIO)\cite{zihao2017event}. However, most methods rely on conventional ANN processing. Spiking Neural Networks (SNNs), which excel in temporal precision, are better suited for handling event-based data. SNNs have been used in control tasks\cite{tang2021deep} but remain underexplored in real-world robotic applications. While multimodal integration of SNNs with traditional sensors has been explored~\cite{yu2023brain}, a unified perception and control framework using SNNs is still lacking.

\begin{figure*}[t!]
	\setlength{\tabcolsep}{1.0pt}
	\centering
	\begin{tabular}{c}
        \includegraphics[width=0.7\textwidth]{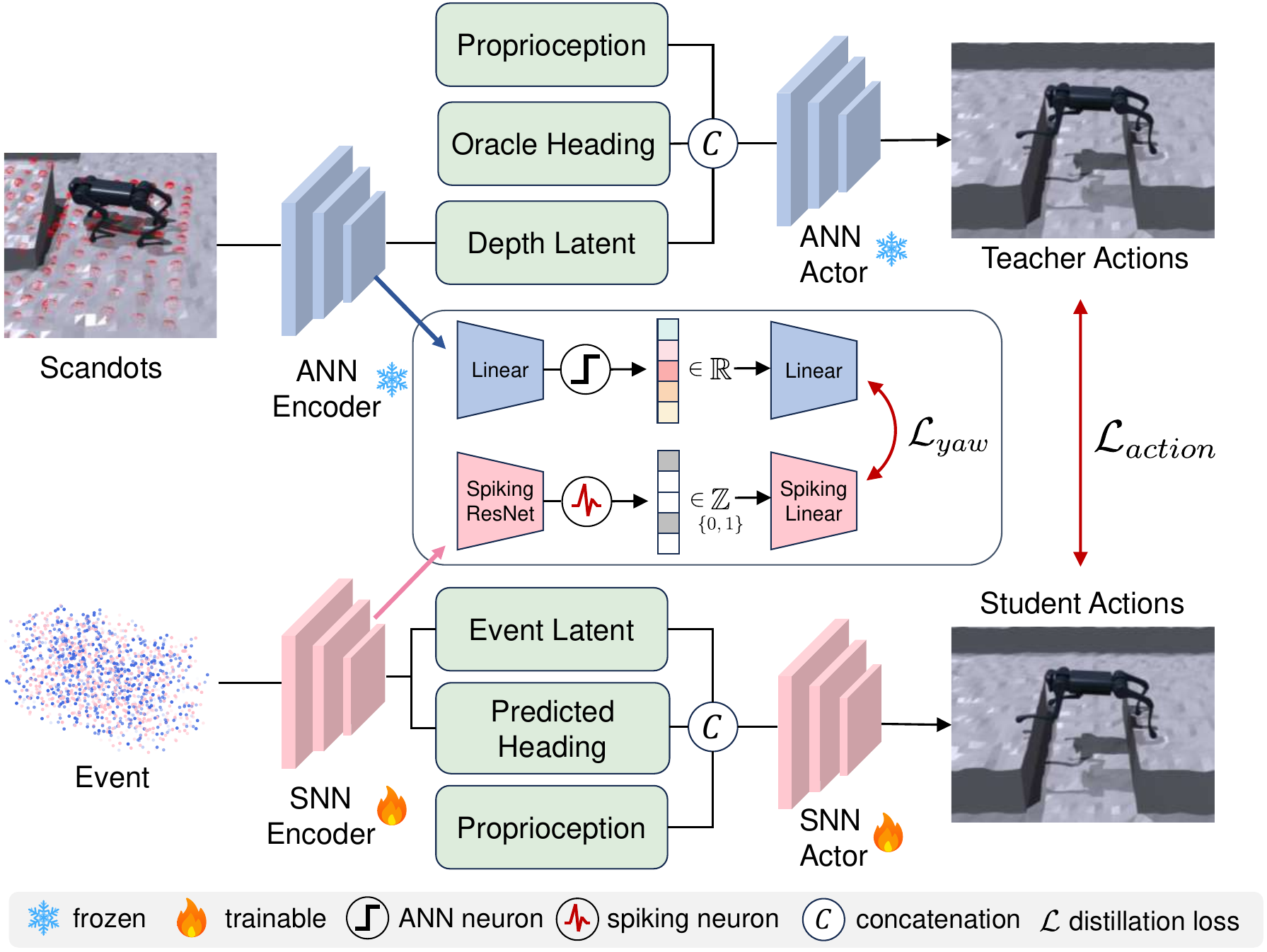} 
	\end{tabular}
	\caption{\textbf{Pipeline of our ES-Parkour ANN-to-SNN distilling process.} Through the distillation process, the extreme parkour capabilities of the ANN are transferred to an SNN, which receives input from an event camera. In the warm-up phase, minimizing the Mean Squared Error (MSE) loss between the outputs of the teacher (ANN) and the student (SNN) networks ensures the student network can closely replicate the teacher network's outputs. Following the warm-up phase, the student network demonstrates basic movement capabilities but encounters challenges with complex terrains. Further interaction and optimization of the student network enhance its performance on complex terrains, closely aligning it with the teacher's performance.
 }
	\label{fig:pipeline}
\end{figure*}

\section{Method}

\subsection{Build Event Camera in Simulation}

Event Camera are bio-inspired sensors, which capture the relative intensity changes asynchronously. In contrast to standard cameras that output 2D images, event cameras output sparse event streams.
When brightness change exceeds a threshold $C$, an event $e_k$ is generated containing position $\textbf{u} = (x,y)$, time $t_k$, and polarity $p_k$:
\begin{equation}
    \Delta L(\textbf{u},t_k) = L(\textbf{u},t_k) - L(\textbf{u},t_k - \Delta t_k) = p_k C.
    \label{eq:event_generate}
\end{equation}

The polarity of an event reflects the direction of the changes. In this paper, we utilize IsaacGym as the simulation and training environment. IsaacGym is a high-performance robotic simulation platform that provides a rich physical simulation environment, enabling us to efficiently train and test quadruped robots in complex scenarios. However, the IsaacGym platform does not natively support the simulation of event cameras.
Therefore, we develop an algorithm to simulate the working principle of event cameras within the IsaacGym environment:

Suppose that in a small time interval, the brightness consistency assumption~\cite{horn1981determining} is conformed, under which the intensity change in a vicinity region remains the same. By using Taylor's expansion, we can approximate intensity change by:
\begin{align}
    \Delta L(\textbf{u}, t) &= L(\textbf{u}, t) - L(\textbf{u}, t-\Delta t), \\
    &= \frac{\delta L}{\delta t}(\textbf{u}, t)\Delta t + O(\Delta t^2)\approx \frac{\delta L}{\delta t}(\textbf{u}, t)\Delta t,
\end{align}
where $\textbf{u}=(x,y)$ denotes the position.
Substituting the brightness constancy assumption ($\frac{\delta L}{\delta t}(\textbf{u}(t),t) + \nabla L(\textbf{u}(t),t) \cdot \textbf{v}(\textbf{u})) = 0.$) into the above equation, we can obtain:
\begin{equation}
    \Delta L(\textbf{u}) \approx -\nabla L(\textbf{u}) \cdot \textbf{v}(\textbf{u})\Delta t, \label{eq:bright}
\end{equation}
which indicates that the brightness changes are caused by intensity gradients $\Delta L = (\frac{\delta L}{\delta x},\frac{\delta L}{\delta y})$
moving with velocity $\textbf{v}(\textbf{u})$ over a displacement $\Delta \textbf{u} =\textbf{v}\Delta t $. With $\textbf{v}(\textbf{u})$ and $\nabla L(\textbf{u}) $, we can get $\Delta L(\textbf{u}) $ to generate event data with Eq.~\ref{eq:event_generate}. In this paper, we adopt the same simulated methods to obtain $\textbf{v}(\textbf{u})$ and $\nabla L(\textbf{u}) $ from \cite{cao2024chasing}, where only a single depth image is required to simulate the corresponding event frames.

Our simulation algorithm calculates pixel changes in the environment in real-time and converts these changes into events that would be output by an event camera,
making the simulated event data as close as possible to the output of a real event camera.

\subsection{Build SNNs in Simulation}
Spiking neural network is a bio-inspired algorithm that mimics the actual signaling process occurring in brains.
Compared to the ANNs, it transmits sparse spikes instead of continuous representations, offering benefits such as low energy consumption and robustness. 
In this paper, we adopt the widely used Leaky Integrate-and-Fire (LIF~\cite{hunsberger2015spiking}) model, which effectively characterizes the dynamic process of spike generation and can be defined as:
\begin{align}
    & V[n] = \beta V[n-1] + \gamma I[n] \label{eq:dis_lif1},\\
    & S[n] = \Theta (V[n] - \vartheta_{\textrm{th}})\label{eq:dis_lif2},
\end{align}
where $n$ is the time step and $\beta$ is the leaky factor that controls the information reserved from the previous time step; $V[n]$ is the membrane potential; $S[n]$ denotes the output spike which equals 1 when there is a spike and 0 otherwise; $\Theta(x)$ is the Heaviside function. When the membrane potential exceeds the threshold $\vartheta_{\textrm{th}}$, the neuron will trigger a spike and resets its membrane potential to $V_{\textrm{reset}}<\vartheta_{\mathrm{th}}$.

\begin{table*}[t]
\centering
\begin{tabularx}{\textwidth}{l|X|X|X}
\toprule
\textbf{Camera Type} & \textbf{RGB Camera} & \textbf{Depth Camera} & \textbf{Event Camera} \\ \hline
Dynamic Range        & Low ($\sim$60dB) & Low       & High ( $\geq$ 120dB )         \\ \hline
Latency              & High  & High       & Low       \\ \hline
Advantages           & $\bullet$ Versatile for many conditions\newline $\bullet$ High color fidelity & $\bullet$ Captures spatial data for 3D modeling\newline $\bullet$ Useful in AR/VR & $\bullet$ Great for capturing movement in high dynamic range scenes \\ \hline
Disadvantages        & $\bullet$ Limited in extreme lighting without HDR & $\bullet$ Limited functionality in diverse light conditions & $\bullet$ Less effective for static scenes \\ \bottomrule
\end{tabularx}
\caption{\textbf{Comparison of RGB, Depth, and Event Cameras.} Event cameras, with their high dynamic range (HDR) and low latency, are ideally suited for robotic applications in outdoor and extreme-exposure environments.}
\label{tab:camera_comparison}
\vspace{-0.2cm}
\end{table*}

\subsection{Learning Process}

\noindent\textbf{Reinforcement Learning on ANN.}
Our policy training framework is structured as a Markov Decision Process (MDP), defined by the tuple $(\mathcal{S},\mathcal{A},\mathcal{R},p,\gamma)$, where $\mathcal{S}$ denotes the state space, $\mathcal{A}$ represents the action space, $\mathcal{R}$ is the reward function, $p$ characterizes the transition probabilities between states for each action-state pair, $\gamma$ is the discount factor applied to rewards. At each time step $t$, the agent receives a state $s_t \in \mathcal{S}$. Based on this observation, the agent selects an action $a_t \in \mathcal{A}$ which is sampled from policy $\pi(a_t|s_t)$. This action leads to a transition $s_t$ to a new state to $s_{t+1}$ determined probabilistically by $s_{t+1} \sim p(s_{t+1}|s_t,a_t)$. And the agent obtains a reward value at each time step $r_t=\mathcal{R}(s_t,a_t)$. The primary goal is to optimize the policy parameters $\theta$ to maximize the reward:
\begin{equation}
    \textnormal{arg}\max_{\theta} \mathbb{E}_{(s_t,a_t) \sim p_\theta(s_t,a_t)} \left[ \sum_{t=0}^{T-1} \gamma^t r_t\right],
\end{equation}
where T denotes the time horizon of MDP. 

Our ANN teacher policy training process follows~\cite{cheng2023extreme} to aim for the policy to not directly learn the skills of traversing difficult terrain, but rather to enable the robot to parkour by learning from rewards and following instructions. Thus, unlike approach~\cite{kumar2021rma}, our method uses privileged information like scandots of terrain, which can be acquired in real-world scenarios, instead of relying on environmental factors like friction. In this phase, the policy externally receives the scandots, and target yaw direction as privileged observations. We utilize various obstacles including gaps, steps, hurdles, and parkour terrain to train the policy. 

\noindent\textbf{Distilling to SNN.}
In the initial phase of our training, we employ ANN to create a model capable of generating action and directional commands for executing parkour tasks with quadruped robots. This foundational step establishes the groundwork for our subsequent transition to a more energy-efficient model. We then embark on a distillation process, where the goal is to train an SNN to emulate the decision-making behavior of the ANN. This process begins with the ANN, serving as the teacher network, and interacting with the simulation environment. We proceed by training the SNN, referred to as the student network, aiming to minimize the Mean Squared Error (MSE) loss between the outputs of both the student and teacher networks. 
To enhance the performance, we further engage the student network in interactions with the environment. During this phase, we continue to measure and minimize the loss 
under identical environmental conditions.
The training process is shown in Figure~\ref{fig:pipeline}, where the distillation loss are defined as:
\begin{align}
\mathcal{L}_{action} &= \frac{1}{n}  \frac{1}{m} \sum_{i=1}^{n} \sum_{j=1}^{m}(action^{\mathrm{ANN}}_{ij} - action^{\mathrm{SNN}}_{ij})^2,\\
\mathcal{L}_{yaw} &= \frac{1}{m} \sum_{j=1}^{m} (yaw^{\mathrm{ANN}}_{j} - yaw^{\mathrm{SNN}}_{j})^2,
\end{align}
where $n$ represents the total number of joints in the quadruped robot and $m$ represents the number of training robots.
This iterative process of fine-tuning and adjustment enables the SNN to closely match the ANN's output patterns across a variety of scenarios. As a result, the original model's performance is preserved, while the computational load and energy consumption are significantly reduced. This efficiency improvement renders the model more suitable for real-time applications on devices with limited power resources, marking the successful completion of our training process.

Due to the wide dynamic range of event cameras, we can distill our SNN model using models trained under normal lighting conditions with depth cameras, achieving the same effect. This means that although our SNN model is trained with external environmental light different from traditional depth cameras, it can still maintain efficient and accurate inference capabilities under extreme lighting conditions (e.g., direct sunlight or low-light scenes). 
Our approach opens up new possibilities for deploying quadruped robots in more challenging environments, enabling them to perform precise perception and rapid response under almost any lighting condition.

\begin{figure}[t]
    \centering    
    \includegraphics[width=0.35\textwidth]{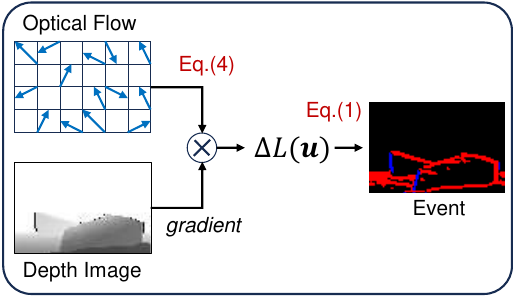} 
    \caption{\textbf{Overview of the event simulation process.} Each depth image can be converted into its corresponding event with the optical flow and image gradient.}
    
    \label{fig:depth2event}
    \vspace{-0.4cm}
\end{figure}

\subsection{Theoretical Energy Consumption Calculation}
\label{subsec:energy}

To calculate the theoretical energy consumption of SNN, we begin by determining the synaptic operations (SOPs). The SOPs for each block in the spiking model can be calculated using the following equation~\cite{zhou2022spikformer}: $\operatorname{SOPs}(l)=fr \times T \times \operatorname{FLOPs}(l)$, where $l$ denotes the block number in the spiking model, $fr$ is the firing rate of the input spike train of the block and $T$ is the time step of the spike neuron. $\operatorname{FLOPs}(l)$ refers to floating point operations of $l$ block. 

To estimate the theoretical energy consumption of our model, we assume that the MAC and AC operations are 32-bit floating-point implementations in $45 nm$ hardware~\cite{horowitz20141},
with energy costs of $E_{MAC} = 4.6 pJ$ and $E_{AC} = 0.9 pJ$, respectively. According to~\cite{panda2020toward, yao2023attention}, the calculation for the theoretical energy consumption of ES-Parkour is given by:
\begin{equation}
 \begin{aligned}
E_{\text {ES-Parkour}} & =E_{MAC} \times \mathrm{FLOP}_{\mathrm{SNN}_\mathrm{Conv}}^1 \\
& +E_{AC} \times\left(\sum_{n=2}^N \mathrm{SOP}_{\mathrm{SNN}_\mathrm{Conv}}^n+\sum_{m=1}^M \mathrm{SOP}_{\mathrm{SNN}_\mathrm{FC}}^m\right)
\end{aligned}   
\end{equation}
where $N$ and $M$ represent the total number of layers of Conv and FC, $E_{MAC}$ and $E_{AC}$ represent the energy cost of MAC and AC operation, $\mathrm{FLOP}_{\mathrm{SNN}_\mathrm{Conv}}$ denotes the FLOPs of the first Conv layer, $\mathrm{SOP}^n_{\mathrm{SNN}_\mathrm{Conv}}$ and $\mathrm{SOP}^m_{\mathrm{SNN}_\mathrm{FC}}$ are the SOPs of $n^{th}$ Conv and $m^{th}$ FC layer, respectively.

\section{Experiments}

\subsection{Training Setting}

During the transition from ANN to SNN in our distillation process, we embark on an extensive training regimen for the student SNN model. This training is conducted within IsaacGym, utilizing a total of 32 parallel robot simulation environments. These environments are specifically chosen to provide a diverse range of challenges and scenarios, thereby ensuring a comprehensive learning experience for the SNN model. To simulate real-world conditions as closely as possible, we sample event images at a frequency of 10Hz, which allows us to capture dynamic changes within the network inferencing effectively. The training process is powered by an NVIDIA 3090 GPU and spans over a duration of 30 hours. In configuring the SNN, we opt for the Integrate-and-Fire (IF) neuron model, renowned for its simplicity and efficiency. The spiking timestep is set as 4, optimizing the balance between responsiveness and computational demand. 

During our training process, we adopt a series of meticulously designed parameters to optimize the performance of our SNN model. Firstly, we set the learning rate to 0.001.
For our encoder network structure, we choose the spiking ResNet-18~\cite{fang2021deep} as our vision backbone.
We also use the GRU module to fuse the latent features encoded from proprioceptive information and event features.
Additionally, we incorporate a 3-layer spiking MLP layer, with sizes [512, 256, 128], to serve as the actor network.

\begin{figure}[t!]
	\setlength{\tabcolsep}{1.0pt}
	\centering
        \includegraphics[width=.4\textwidth]{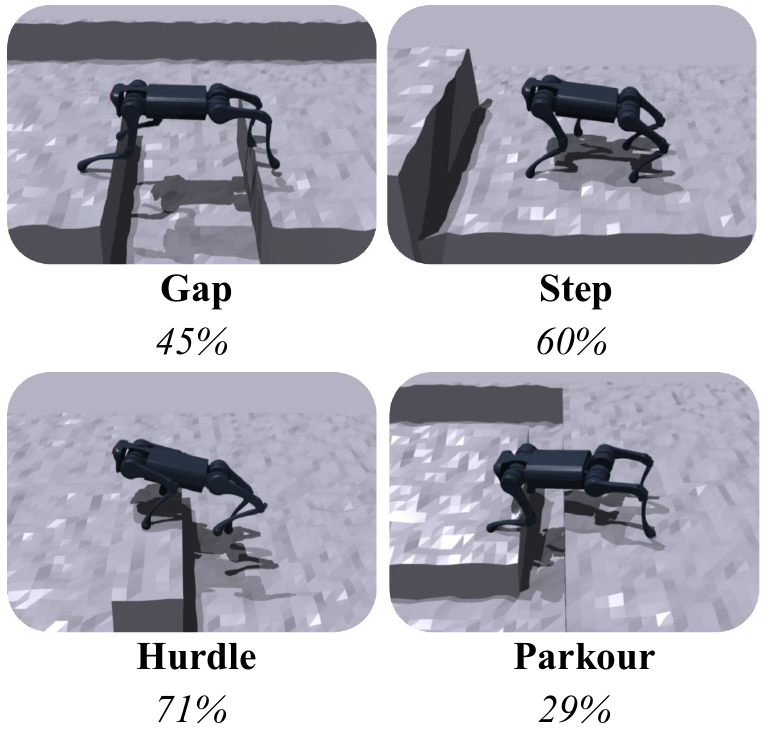} 
	\caption{We evaluate our SNN strategy across four different scenarios. The figure shows the shapes related to each. The top row indicates the type of terrain, while the bottom row displays the success rate for each situation.}
	\label{fig:scenarios}
\end{figure}

\subsection{Simulation Results}

During the training phase of our quadruped robot's Spiking Neural Network, we closely monitor the terrain level curve to assess the robot's ability to adapt to complex terrains. We evaluate our SNN strategy and the results are shown in Figure~\ref{fig:scenarios}. 
This method gradually guides the robot to face tasks of increasing difficulty, significantly enhancing its adaptability and performance under various environmental conditions. It is also important to note that our training achievements are made under varying lighting conditions, which means our curriculum learning can robustly handle changes in lighting.

\begin{table}[t]
        \centering
        \renewcommand\arraystretch{1.18}
        \resizebox{\linewidth}{!}{%
        \begin{tabular}{ccccc}
            \toprule
            \multirow{2}{*}{Encoder Type} & \multicolumn{2}{c}{SNN} & \multicolumn{1}{c}{ANN} & \multicolumn{1}{c}{Efficiency $\downarrow$}\\
            \cmidrule(lr){2-3} \cmidrule(lr){4-4} \cmidrule(lr){5-5}
             & FLOPs & SOPs & FLOPs & OPs(SNN):OPs(ANN)\\
            \midrule
             \multicolumn{1}{c}{ResNet} & 8.00$\times$e$^{6}$  & 8.76$\times$e$^{7}$ & 2.04$\times$e$^{8}$ & \textbf{0.46}~:~1 \\
             \multicolumn{1}{c}{MLP} & 7.17$\times$e$^{6}$  & 2.61$\times$e$^{6}$ & 3.31$\times$e$^{7}$ & \textbf{0.29}~:~1\\
             \bottomrule
        \end{tabular}%
        }
        \captionof{table}{Comparisons of the number of operations (FLOPs/SOPs) between the vision encoder of Parkour and ES-Parkour. SNN yields lower operating times than its ANN counterpart.}
        \label{tab:operations_results}
\end{table}
    
\begin{table}[t]
    \centering
    \resizebox{\linewidth}{!}{%
    \begin{tabular}{cccc}
        \toprule
        \multirow{2}{*}{Module} & \multicolumn{2}{c}{Encoder} & \multicolumn{1}{c}{Actor} \\
        \cmidrule(lr){2-3} \cmidrule(lr){4-4}
         & ResNet (11.19M) & MLP (8.01M) & MLP (0.26M)\\
        \midrule
         \multicolumn{1}{c}{ANN Power (mJ)} & 0.94  & 0.15 & 1.08e$^{-3}$ \\
         \multicolumn{1}{c}{SNN Power (mJ)} & \textbf{0.11} & \textbf{0.04} & \textbf{3.30e$^{-4}$} \\
         \multicolumn{1}{c}{Energy Saving} & 88.29\% & 73.33\% & 69.44\%\\
         \bottomrule
    \end{tabular}%
    }
    \captionof{table}{Comparisons of Energy Consumption between origin Parkour (ANN model) and ES-Parkour (SNN model). Our ES-Parkour achieves extreme energy saving (up to \textbf{88.29\%}) in each module.}
    \label{tab:energy_results}
    \vspace{-0.3cm}
\end{table}

\subsection{Analysis of Computing Efficiency}
\subsubsection{Comparisons of the number of operations.}
Given that the majority of computational demands in neural networks stem from matrix operations, this section explores the analysis and comparison of the operational counts within the visual encoders of both ANN and SNN. This comparison aims to validate the efficiency of our ES-Parkour system. According to Table~\ref{tab:operations_results}, the SNN consistently demonstrates a lower total operational count (including both FLOPs and SOPs) compared to the ANN, regardless of whether ResNet or MLP serves as the visual backbone.
This difference arises because, the non-spiking portion of the feature (i.e., zero value) in SNNs does not consume computational resources during matrix operations. As a result, the overall number of operations for SNNs significantly falls below that of ANNs. We define an operational efficiency metric:
\begin{equation}
    \text{Efficiency} = \frac{\mathrm{OPs}(\mathrm{SNN})}{\mathrm{OPs}(\mathrm{ANN})} = \frac{\mathrm{FLOP}_{\mathrm{SNN}} + \mathrm{SOP}_{\mathrm{SNN}}}{\mathrm{FLOP}_{\mathrm{ANN}}},
\end{equation}
where this metric measures the relative energy efficiency ratio, a lower efficiency value (i.e., less than 1) indicates a higher energy efficiency of SNNs compared to ANNs. The reduced efficiency values presented in the table underline the computational efficiency of our ES-Parkour system.

\begin{table}[t]
    \centering
    \renewcommand\arraystretch{1.2}
    \vspace{-1mm}
\scalebox{1.12}{
\begin{tabular}{ccccc}
\hline
     & \multicolumn{1}{c}{Gap} & \multicolumn{1}{c}{Step} & \multicolumn{1}{c}{Hurdle} & \multicolumn{1}{c}{Parkour}  \\ \hline
ANN      & 808 16                                & 876.23                                 & 853.32                              & 1008.6                                                         \\
SNN (ours)       & 813.45                               & 869.27                                & 862.01                              & 997.54  \\                                                     
\hline
\end{tabular}}
\caption{Comparisons of the average robots' joints motor energy (mJ) between ANN and SNN.}
\label{energy_table}
\vspace{-3mm}
\end{table}

\subsubsection{Evaluation of the energy consumption.}
To further emphasize the low-energy nature of our ES-Parkour, we conduct a detailed comparative analysis of the energy consumption between the proposed ES-Parkour and its corresponding ANN model. As shown in Table~\ref{tab:energy_results}, using the ResNet scenario as an example, our ES-Parkour presents significantly lower energy consumption, amounting to merely 11.7\% of that exhibited by the ANN model. This corresponds to an extreme energy-saving (88.3\%) by utilizing SNN. Moreover, the actor module of the ES-Parkour further exemplifies energy conservation compared with ANN Parkour, with 3.30$\times$e$^{-4}$ vs. 1.08$\times$e$^{-3}$, demonstrating the superior low-energy benefits of our systems. We can maintain the same power consumption at the joint level as with ANN, but with better environmental adaptability and lower computational burden as shown in Table~\ref{energy_table}. Our extensive testing has validated the overall feasibility and performance advantages of the system.

\begin{table}[t]
    \centering 
\scalebox{0.85}{\begin{tabular}{ccccc}
\toprule
Scenarios                & normal-light & overexposed & underexposed & \multicolumn{1}{l}{high-speed} \\ \midrule
Anymal parkour~\cite{hoeller2023anymal}  & $\checkmark$            & $\checkmark$       & $\checkmark$    & $\times$                              \\
Extreme parkour~\cite{cheng2023extreme} & $\checkmark$            & $\times$       & $\times$    & $\times$                              \\
Robot parkour~\cite{zhuang2023robot}   & $\checkmark$            & $\times$       & $\times$    & $\times$                              \\
\textbf{ES-Parkour (ours)}           & $\checkmark$            & $\checkmark$       & $\checkmark$    & $\checkmark$                              \\ \bottomrule
\end{tabular}}

    \caption{Comparison of the abilities of different methods in extreme scenarios.}
    \vspace{-0.4cm}
    \label{table: Comparison}
\end{table}
\section{Conclusion}
In this paper, by integrating Spiking Neural Networks (SNN) and event cameras, we not only address the challenges of power consumption and computational load inherent in traditional deep learning models for quadruped robot parkour but also forge a new pathway for enhancing robot perception and control. This approach enables more efficient and adaptive responses in complex environments. We compare the work of robot parkour in Table~\ref{table: Comparison}, and our work is the only one that can be tested under all environmental conditions. 

Due to the difficulty in obtaining SNN chips, our work has not been tested on actual robots. However, as with previous robotic validation efforts, we have extensively tested our system in simulations to ensure its feasibility on actual robots. This phased research ensures the sustainability of our study. In the future, we will continue to refine our system and advance the integration of the SNN chips with actual robots.

\bibliographystyle{IEEEbib}me
\bibliography{icme}

\end{document}